\documentclass[11pt,a4paper]{article}
\usepackage{authblk}
\usepackage[hyperref]{emnlp2020}
\usepackage{times}
\usepackage{latexsym}

\usepackage{microtype}
\usepackage{times}
\usepackage{latexsym}
\usepackage[symbol]{footmisc}
\usepackage{adjustbox}
\usepackage{enumitem}
\usepackage{graphicx}
\usepackage{caption}
\usepackage{subcaption}
\graphicspath{ {./figs/} }
\usepackage{tabularx}
\usepackage{float}

\aclfinalcopy 
\def\aclpaperid{2491} 

\title{On Losses for Modern Language Models}

\author[1,2]{Stéphane Aroca-Ouellette}
\author[1,2,3]{Frank Rudzicz}
\affil[1]{University of Toronto}
\affil[2]{Vector Institute for Artificial Intelligence}
\affil[3]{Li Ka Shing Knowledge Institute, St Michael's Hospital}
\affil[ ]{\texttt{\{stephaneao, frank\}@cs.toronto.edu}}

\date{June 1st 2020}

\begin{document}
\maketitle
\begin{abstract}
BERT set many state-of-the-art results over varied NLU benchmarks by pre-training over two tasks: masked language modelling (MLM) and next sentence prediction (NSP), the latter of which has been highly criticized. In this paper, we 1) clarify NSP's effect on BERT pre-training, 2) explore fourteen possible auxiliary pre-training tasks, of which seven are novel to modern language models, and 3) investigate different ways to include multiple tasks into pre-training. We show that NSP is detrimental to training due to its context splitting and shallow semantic signal. We also identify six auxiliary pre-training tasks -- sentence ordering, adjacent sentence prediction, TF prediction, TF-IDF prediction, a FastSent variant, and a Quick Thoughts variant -- that outperform a pure MLM baseline. Finally, we demonstrate that using multiple tasks in a multi-task pre-training framework provides better results than using any single auxiliary task. Using these methods, we outperform BERT\textsubscript{Base} on the GLUE benchmark using fewer than a quarter of the training tokens.
\end{abstract}

\section{Introduction}

When \citet{bert} released BERT, a transformer network \citep{transformer} trained using a `masked language model' (MLM) task and a `next sentence prediction' (NSP), it redefined the NLP landscape, establishing itself as the state-of-the-art (SoTA) on many natural language understanding (NLU) benchmarks including the GLUE \citep{glue}, SQuAD \citep{squad}, and SWAG \citep{swag} benchmarks.

Many models inspired by BERT have since surpassed its performance. However, in contrast to the original BERT paper, many obtained better results by excluding the NSP task. Some, such as XLNET \citep{xlnet} and RoBERTa \citep{roberta}, rely solely on a MLM variant, while others \citep{structBERT, spanbert, psp, ernie2} incorporate one or more different auxiliary loss functions. To our knowledge, there is no published work comparing or fully exploring auxiliary tasks for modern language models.

With multi-task learning's long history in transfer learning \citep{multitask, actormimic, meta}, its use in language understanding models deserves further exploration. In this paper, we study existing and novel auxiliary tasks in a BERT paradigm to guide future research in an informed manner. Specifically, we test and provide insight on: 1) NSP's effect on BERT pre-training; 2) the result of 14 other auxiliary tasks on BERT pre-training; 3) how to combine multiple tasks in BERT pre-training; and 4) the advantages of multi-task learning in BERT pre-training. Although all experiments in this paper are conducted using BERT, we believe the results are applicable to BERT's successors (e.g. XLNET, RoBERTa, ERNIE...) and future models. The code is available at \url{https://github.com/StephAO/olfmlm}.

\section{Related work}
As with most deep learning, language representations require large datasets. While there exists corpora of labelled text, the vast majority of language data exists as raw, unlabelled text. Accordingly, many language embedding methods, and all those described below, rely solely on unsupervised or self-supervised tasks. 

\subsection{Pre-transformer sentence embeddings}
Skip-Thoughts \citep{skipthoughts} was the first deep learning sentence embedding model. Its training objective, inspired by word2vec \citep{word2vec}, used RNNs to reconstruct the previous and next sentence from a given sentence. Like word2vec, similar sentences shared similar embeddings, and while it exhibited promising results, it was slow to train due to its encoding and double decoding of sentences through RNNs. \citet{sdae}'s FastSent tried to follow the same sequential sentence paradigm at a reduced training cost by encoding a sentence using a bag-of-words approach and maximizing the probability of words in adjacent sentences. Later, Quick Thoughts \citep{quickthoughts} managed to maintain the sequential sentences objective while supporting ordered words. Using two RNN models, $f(s)$ and $g(s)$, they embedded a first set of sentences using $f(s)$ and a second set consisting of the subsequent sentences using $g(s)$. They jointly train the two models to predict the consecutive sentences from a set of candidates by comparing inner products. This resembles a referential game \citep{lewis} where $f(s)$ and $g(s)$ are the sender and receiver respectively. 

The previous methods relied on the premise that adjacent text is semantically similar, but other sentence embedding methods have relied on other linguistic properties. The sequential denoising autoencoder (SDAE) \citep{sdae} corrupts a sentence through deletion or word order swapping, encodes it, then attempts to decode the original uncorrupted sentence from the encoding. This shares the same underlying sequence-consistency concept as BERT's MLM. \citet{corrupt} also focused on sequence consistency by predicting whether or not a sentence had been corrupted through deletion, insertion, replacement, or permutation. These methods only require individual sentences rather than a set of sequential sentences.

\subsection{Transformer-based sentence embeddings}
The development of the transformer network \citep{transformer} overcame the sequential bottleneck of RNNs by fully utilizing the parallelization of modern processing units, and enabling language models to train on significantly more data in less time. GPT \citep{gpt} and its successor GPT-2 \citep{gpt2} were the first models to fully leverage this breakthrough. Following traditional language modelling, their training objective is to maximize the probability of a sequence of tokens $x$ using the products of their conditional probabilities  $p(x) = \prod_{i=1}^{n} (t_n\, |\, t_{n-1}, ..., t_1)$. 

\citet{bert} addressed the limitation of unidirectional context in traditional language modelling in their development of Bidirectional Encoder Representations from Transformers (BERT) -- a transformer trained using a masked language modelling (MLM) task and next-sentence prediction (NSP) task on approximately 137 billion tokens from a 3.3 billion word corpus created from the concatenation the BooksCorpus \citep{bookcorpus} and English Wikipedia datasets. The masked language model modifies the traditional language model to consider the bidirectional context in its prediction. For each sequence, 15\% of tokens are replaced with a [MASK] token. The model is then trained to predict the masked words. The NSP task uses the output embedding of the [CLS] token that prepends the sequence to predict whether the second sentence follows the first or is from a different document. BERT's original paper claimed that this task improved performance on downstream natural language inference (NLI) tasks. 

MASS \citep{mass}, ERNIE \citep{ernie}, and SpanBERT \citep{spanbert} extended the MLM task by masking a sequence of contiguous tokens instead of a single token. All three demonstrated the superiority of this approach. MASS used a seq2seq model \citep{seq2seq} to decode the sequence of masked tokens. ERNIE used larger sequences of tokens over the course of three stages -- first identical to BERT, then masking phrases, then masking full entities. They additionally added a dialogue language model task using the CLS token to classify between question-response pairs and random pairs. SpanBERT uses spans of sampled lengths and a `span boundary objective' where the token embeddings adjacent to the span are used to predict the masked span. Each of their additions provided gains on a range of downstream tasks, with maximal gains using both. They additionally showed that NSP is detrimental to training, hypothesizing that the context splitting required for NSP is more detrimental than the advantages provided from the task. \citet{psp} argued that NSP is semantically shallow and does not leverage BERT'S bidirectional nature, and replaced NSP with a three-way classification task of identifying whether one sentence follows or precedes another, or is from a different document. Using this simple change, they achieved a modest improvement over the BERT baseline. 

XLNET \citep{xlnet} used permuted sentences to combine the true language modelling objective of GPT-2  \citep{gpt2} and BERT's insight of bi-directional context. It included the advancements from transformer-XL \citep{transformerxl} to increase the context length, and created a larger training dataset. It also ran a small ablation study and found that removing the NSP task improved  overall results. XLNET beat BERT on 20 tasks, achieving SoTA on 18. Shortly thereafter, \citet{roberta} introduced RoBERTa, which followed the core concepts of BERT closely, but optimized design choices, such as dynamically masking tokens each epoch instead of pre-masking the entire dataset, increasing the batch size, using full sentences in each batch, and removing the NSP loss. With these changes, and an increased dataset, they matched XLNET's performance. 

\citet{ernie2}'s ERNIE 2.0 made further gains in the GLUE leaderboard\footnote[1]{\url{https://gluebenchmark.com/leaderboard}} by incrementally adding seven tasks in a ``continual multi-task learning" framework. They trained on ERNIE's original token/phrase/entity masking, capitalization prediction, token-document prediction, sentence re-ordering, sentence distance prediction, discourse relations, and information retrieval (IR) relevance. They provided no information on the benefit from any of the individual tasks or the ordering of the tasks. \citet{t5}'s T5, also high on the leaderboard, achieved their results using an encoder-decoder variant of BERT. Through rigorous experimentation of implementation details, culminating in a gigantic 11 billion parameter model trained on more data, using spans, multi-task learning on the supervised downstream tasks, and using beam search, they achieved SoTA on a vast array of tasks. \citet{structBERT}'s StructBERT used a word structural objective, where the model has to recover a shuffled tri-gram, and sentence structural objective -- identical to \citet{psp}'s three-way classification task, to place high on the leaderboard as well.

\section{Method}

\subsection{Baselines}
Our primary motivation in this paper is to study and survey auxiliary pre-training tasks for multi-task learning for modern language understanding models. In this case, `modern' is a transformer-based model pre-trained on a large unlabelled corpus using a form of masked language modelling. This definition captures the large majority of recently successful language models. For our baseline we choose BERT, as it is the basis for subsequent models, but only include the MLM task as the benefits of the NSP task are debated \citep{roberta,xlnet}. For computational reasons we use BERT\textsubscript{Base} ($L=12$, $H=768$, $A=12$, Total Parameters=110M), and use the uncased WordPiece tokenizer \citep{wordpiece} with vocabulary size of 30522 provided by Google\footnote[2]{\url{https://github.com/google-research/bert}}.

\subsection{Auxiliary pre-training tasks}
\label{sec:tasks}
To provide a fair comparison and due to computational constraints, we limit the scope of our investigation to auxiliary tasks that can be directly used on any corpus of unlabelled data, do not require any language decoding, and require at most one additional classification layer. This excludes the discourse relation task, the IR relevance task \citep{ernie2}, and the dialogue language modelling task \citep{ernie} as they require datasets that violate these constraints. We also exclude a Skip-Thoughts approach as sequentially decoding outputs would require significantly more computational resources. Token level tasks only use token embeddings as input. Sentence-level tasks use the [CLS] token embedding as input. The FastSent variant uses both, but we label it as a sentence-level task as it does require the sentence embedding (from the [CLS] token). Tasks that have not previously been applied to modern language models are italicized. We investigate the following tasks:  
\newline\newline
\noindent \textbf{Token level tasks}
\begin{enumerate}
\setlength{\itemsep}{0mm}
  \item \textit{Term Frequency prediction (TF)}: Regression predicting a token's frequency in the rest of the document. The frequency is re-scaled between 0 and 10 per document.
  \item \textit{Term Frequency-Inverse Document Frequency prediction (TF-IDF)}: Regression predicting a token's tf-idf that has been re-scaled between 0 and 10 per document.
  \item Sentence Boundary Objective (SBO): Predict the masked token given the embeddings of the adjacent tokens.
  \item Trigram-Shuffling (TGS): 6-way classification predicting the original order of shuffled tri-grams.
  \item \textit{Token Corruption Prediction (TCP)}: Binary classification of whether a token has been corrupted (inserted, replaced, permuted) or not.
  \item Capitalization Prediction (Cap.): Binary, whether a token is capitalized or not.
  \item \textit{Token Length Prediction (TLP)}: Regression to predict the length of the WordPiece token.
\end{enumerate}
\textbf{Sentence level tasks}
\begin{enumerate}
\setlength{\itemsep}{0mm}
  \setcounter{enumi}{7}
  \item Next Sentence Prediction (NSP): Binary, whether the second sentence follows the first or comes from a separate document.
  \item Adjacent Sentence Prediction (ASP): 3-way classification whether the second sentence proceeds the first, precedes the first, or they come from separate documents. 
  \item Sentence Ordering (SO): Binary, predicting if the two sentences are in or out of order.
  \item Sentence Distance Prediction (SDP): 3-way classification of whether the second sentence proceeds, the two sentences are non-contiguous from the same document, or come from separate documents.  
  \item \textit{Sentence Corruption Prediction (SCP)}: Binary classification of whether a tokens in a sentence have been corrupted (inserted, replaced, permuted) or not.
  \item \textit{Quick Thoughts variant (QT)}: Split each batch into two, where the second half contains the subsequent sentences of the first half (e.g. with batch size 32, sentence 17 follows sentence 1, sentence 18 follows sentence 2,...). We use an energy-based model to predict the correct continuation for each sentence in the first half where the energy between two sentences is defined by the negative cosine similarity of their [CLS] embeddings. We use one model to encode both halves concurrently. See Figure \ref{figs:final_model}.
  \item \textit{FastSent variant (FS)}: Split each batch into two, where the second half contains the subsequent sentences of the first half (same as QT above). The loss is defined as cross-entropy between 1.0 and the cosine similarity of a sentence [CLS] embedding and the other sentence token embeddings ([CLS] embedding from the first half with token embeddings from the second half and [CLS] embeddings from second half with token embeddigns from the first half). We use one model to encode both halves concurrently. 
\end{enumerate}

\subsection{Combining tasks}
\label{sec:combinations}
BERT originally proposed summing the MLM and NSP losses directly. ERNIE uses significantly more losses and proposes a continual multi-task learning framework to incorporate them, in which they incrementally add new tasks while sampling previously learnt tasks. To provide insight on how best to combine tasks, we investigate the six following ways of combining a set of tasks for BERT pre-training:
\setlist{nolistsep}
\begin{enumerate}[noitemsep]
    \item Sum losses from all tasks (sum.)
    \item Incrementally add tasks, summing the losses from all added tasks (Inc.)
    \item Alternating between tasks each iteration (Alt.)
    \item Alternating between auxiliary tasks each iteration and summing it with MLM (Alt.+)
    \item ERNIE's continual multi-task learning (CMTL), for more detail see Appendix \ref{sec:cmtl}
    \item ERNIE's continual multi-task learning on auxiliary tasks summed with MLM (CMTL+)
\end{enumerate}
We note that both a direct summation and a simple incremental approach cannot accommodate tasks that require different input structures -- for example sentence ordering, which requires that the two sentences are always adjacent, cannot be trained simultaneously with next sentence prediction, which requires sentences from different documents at times -- or different corpora, such as ERNIE 2.0's IR relevance dataset.

\begin{table*}[h!]
\centering
\small
\begin{tabular}{|l|c|c|c|c|c|c|c|c|c|c|}
\hline
Aux. Task & \begin{tabular}[c]{@{}c@{}}MNLI \\ 392k\end{tabular} & \begin{tabular}[c]{@{}c@{}} QQP \\ 363k\end{tabular} & \begin{tabular}[c]{@{}c@{}}QNLI \\ 108k\end{tabular} & \begin{tabular}[c]{@{}c@{}}SST-2\\ 67k\end{tabular} & \begin{tabular}[c]{@{}c@{}}CoLA \\ 8.5k\end{tabular} & \begin{tabular}[c]{@{}c@{}}STS-B\\ 5.7k\end{tabular} & \begin{tabular}[c]{@{}c@{}}MRPC \\ 3.5K\end{tabular} & \begin{tabular}[c]{@{}c@{}} RTE \\ 2.5k\end{tabular} & \begin{tabular}[c]{@{}c@{}}Avg. \\ -\end{tabular} \\ \hline
None (MLM) & 80.3 & 88.0 & 86.7 & 91.6 & 51.5 & 84.9 & 86.9 & 60.6 & 78.8 \\ \hline
TF & 81.5 & \underline{88.7} & 89.1 & 90.9 & 46.8 & 87.0 & 87.3 & 62.1 & 79.2 \\
TF-IDF & 81.2 & 88.6 & \underline{89.4} & 90.5 & 46.7 & 86.8 & \underline{88.8} & 63.2 & \textbf{79.4} \\
SBO & 80.5 & 88.0 & 89.1 & \underline{92.5} & 48.8 & 85.4 & 86.6 & 56.0 & 78.4 \\
TGS & 80.5 & 88.2 & 87.1 & 90.6 & 50.3 & 85.4 & 87.8 & 58.1 & 78.5 \\
TCP & 81.3 & 88.5 & 88.0 & 91.5 & 49.7 & 85.7 & 87.0 & 58.1 & 78.7 \\
Cap. & 81.1 & 88.6 & 87.0 & 91.3 & 48.0 & 85.8 & 86.0 & 57.8 & 78.2 \\
TLP & 80.8 & 88.3 & 87.7 & 91.5 & 47.0 & 86.0 & 86.1 & 59.6 & 78.4 \\
\hline
NSP & 79.9 & 87.1 & 86.0 & 90.9 & 48.3 & 84.0 & 85.4 & 58.1 & 77.5 \\
ASP & 80.4 & 88.4 & 88.9 & 89.9 & 42.2 & 86.9 & 87.3 & \underline{68.2} & 79.0 \\
SO & 80.9 & 88.6 & 89.2 & 89.8 & 44.1 & \underline{87.4} & 86.4 & 66.1 & 79.1 \\
SDP & 79.9 & 87.9 & 87.8 & 90.3 & 47.7 & 85.9 & 87.7 & 62.5 & 78.7 \\
QT & 81.6 & 88.6 & 88.7 & 91.4 & \underline{55.6} & 86.2 & 87.1 & 63.5 & \textbf{\underline{80.3}} \\
FS & \underline{81.9} & 88.6 & 88.4 & 91.5 & 55.1 & 86.6 & 88.3 & 59.2 & \textbf{80.0} \\
SCP & 80.4 & 88.4 & 87.6 & 90.4 & 46.6 & 85.3 & 86.4 & 59.2 & 78.0 \\\hline

\end{tabular}
\caption{Test results on GLUE development set for models pre-trained on MLM (No Aux.) and MLM + auxiliary tasks trained over 10 billion tokens. F1 scores are reported for QQP and MRPC, Spearman correlations are reported for STS-B, and
accuracy scores are reported for the other tasks. Refer to section \ref{sec:tasks} for a description of each task. Best results in each column are underlined. Averages above two estimated $\sigma$s of the MLM baseline are bolded.}
\label{table:aux_tasks}
\end{table*}

\begin{table*}[t]
\centering
\small
\begin{tabular}{|l|c|c|c|c|c|c|c|c|c|c|}
\hline
 & \begin{tabular}[c]{@{}c@{}}MNLI \\ 392k\end{tabular} & \begin{tabular}[c]{@{}c@{}} QQP \\ 363k\end{tabular} & \begin{tabular}[c]{@{}c@{}}QNLI \\ 108k\end{tabular} & \begin{tabular}[c]{@{}c@{}}SST-2\\ 67k\end{tabular} & \begin{tabular}[c]{@{}c@{}}CoLA \\ 8.5k\end{tabular} & \begin{tabular}[c]{@{}c@{}}STS-B\\ 5.7k\end{tabular} & \begin{tabular}[c]{@{}c@{}}MRPC \\ 3.5K\end{tabular} & \begin{tabular}[c]{@{}c@{}} RTE \\ 2.5k\end{tabular} & \begin{tabular}[c]{@{}c@{}}Avg. \\ -\end{tabular} \\ \hline\hline
MLM & 80.3 & 88.0 & 86.7 & \underline{91.6} & 51.5 & 84.9 & 86.9 & 60.6 & 78.8 \\
QT & 81.6 & 88.6 & 88.7 & 91.4 & \underline{55.6} & 86.2 & 87.1 & 63.5 & \textbf{80.3} \\\hline\hline
Sum. & \underline{82.0} & \underline{89.0} & \underline{90.5} & 91.2 & 49.4 & 88.3 & 89.1 & \underline{70.8} & \textbf{81.4} \\
Inc. & 80.9 & 88.8 & 89.6 & 90.8 & 50.6 & 87.6 & 86.3 & 69.3 & \textbf{80.6} \\\hline
Alt. & 79.8 & 88.4 & 89.3 & 89.3 & 44.3 & 86.8 & 86.2 & 70.4 & \textbf{79.4} \\
Alt.+ & 81.5 & \underline{89.0} & 90.1 & 90.6 & 55.3 & 87.9 & 87.0 & 68.6 & \textbf{81.3} \\\hline
CMTL & 79.6 & 88.2 & 88.8 & 89.7 & 40.3 & 87.1 & 86.1 & 66.8 & 78.4 \\
CMTL+ & 81.7 & 88.6 & 90.3 & 91.3 & 53.9 & \underline{88.5} & \underline{89.2} & 70.4 & \textbf{\underline{81.7}} \\\hline
\end{tabular}
\caption{Results on GLUE development set for models pre-trained on MLM (our baseline), MLM + QT (best single auxiliary task model) and different combinations of the best performing tasks. Refer to section \ref{sec:combinations} for more detail. Best results in each column are underlined. Averages above two estimated $\sigma$s of the MLM baseline are bolded.}
\label{table:comb_tasks}
\end{table*}

\subsection{Input Representation}
To construct the input embedding to the transformer, we sum token embeddings, learned position embeddings, learned sentence type (sentence A or B) embeddings, and, to enable ERNIE's continual multi-task learning, a learned task id embeddings .

\subsection{Dataset}
We follow precedent in using the BookCorpus\footnote[3]{Unfortunately, the BookCorpus is no longer publicly available.} \citep{bookcorpus} and Wikipedia dataset as our corpora. We filter the Wikipedia corpus in the same fashion as BERT, ignoring lists, tables, and headers. We additionally filter documents that have: fewer than 10 words or fewer than 4 sentences. This excludes small uninformative documents. We additionally segment long documents into documents of roughly 1024 tokens. This creates a corpus with 2.7 billion words (3.8 billion tokens) divided into 6.8 million documents.

\subsection{Pre-Training Details}
For all tests, we train on 10 billion tokens using an Adam optimizer \citep{adam} with a learning rate of 1e-4 that warms-up over the first 1\% of tokens and linearly decays after, batch size = 128, max sequence length = 128, $\beta_1$ = 0.9, $\beta_2$ = 0.999, L2 weight decay of 0.01, and a dropout probability of 0.1. In accordance with other papers, we use a gelu activation \citep{gelu}. Using four p100 GPUs, it takes between 13 and 15 hours to train a model for each one billion token epoch depending on the tasks used.

\subsection{Fine-Tuning Details}
All models are tested on the GLUE \citep{glue} benchmark, as it has been accepted by the community as a benchmark for NLU. We also compare the final best model and our baseline on the SuperGLUE \citep{superglue} benchmark. Following \citet{bert, psp}, we disregard GLUE's problematic WNLI task. Due to GLUE's private test set, and the number of experiments performed, the results are on the available development set except for the final results in Tables \ref{table:glue} and \ref{table:superglue}. To fine-tune the model on the GLUE dataset, we use Jiant's \citep{jiant} provided code\footnote[4]{https://github.com/nyu-mll/jiant}. We limit the maximum number of epochs to 3 and we run the fine-tuning procedure three times with learning rates = 5e-5, 3e-5, 2e-5 and take the best results for each task individually across these runs. This is done to reduce the variance in the results that comes from the low-resource tasks CoLA, RTE, and MRPC. For all other fine-tuning parameters, we use the default values provided by jiant unless otherwise stated.

\subsection{Final Model}
\label{sec:final_model}
\begin{figure*}[t]
\begin{adjustbox}{width=0.9\textwidth}
    \centering
    \includegraphics[width=0.9\textwidth]{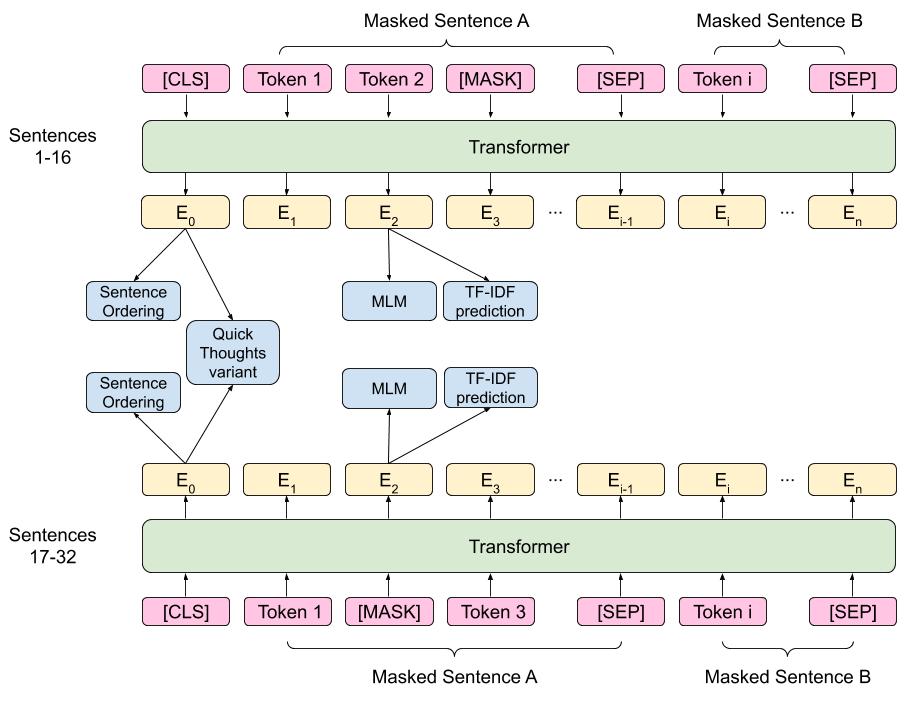}
\end{adjustbox}
\caption{Architecture used for the combined tasks tests. Sentences 17-32 are the continuations of sentences 1-16 respectively (1-17, 2-18, 3-19...). The two halves of the batch are only split for clarity of the Quick Thoughts variant task; they are embedded at the same time by the same network. Though only depicted on only one token, the token level tasks (MLM, TF-IDF prediction) are trained across all token embeddings.}
\label{figs:final_model}
\end{figure*}

Our final CMLT+ model is shown in Figure \ref{figs:final_model} to help visualize the inputs to each task.

\section{Results}
In this section, we present the results from an array of different tests. Due to the stochastic nature of the training, we would ideally run each test numerous times. However, this is prohibitively expensive due to the computational costs. Building from \citet{t5}'s experimental approach, we instead calculate the standard deviation for 5 independent trainings of the baseline MLM-only model, the MLM + NSP model, and our CMTL+ model. We find $\sigma_{MLM}=0.198$, $\sigma_{NSP}=0.222$, $\sigma_{CMTL+}=0.273$, and use the highest, $\sigma={0.273}$, as an estimate for the standard deviation across all experiments. See Appendix \ref{sec:sig test} for more detail. This is comparable to \citet{t5}'s estimated standard deviation of 0.235. In each table, we boldface all average GLUE scores that are two estimated standard deviations above the MLM baseline. For the average GLUE score, we follow \citet{glue} and average the macro averages of each task. This is different than averaging the numbers in a row as we only report one metric per task.

\begin{figure*}[!t]
\begin{adjustbox}{width=1\textwidth}
\centering
\includegraphics[width=1.0\linewidth]{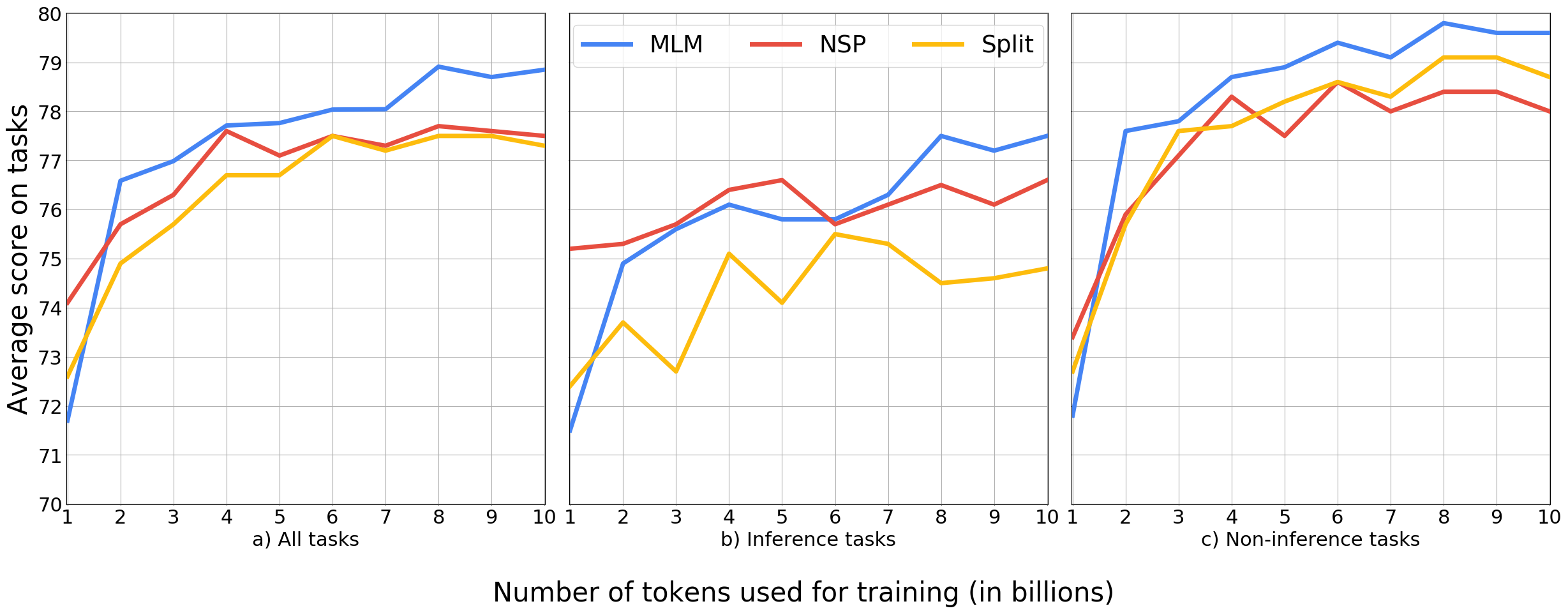}
\end{adjustbox}
\caption{Average results on a) all GLUE tasks, b) only inference tasks (MNLI, QNLI, RTE), and c) non-inference tasks (QQP,  SST-2,  CoLA,  STS-B,  MRPC) for models trained on MLM, MLM + NSP, and MLM with NSP's split context but no NSP loss (Split) throughout training over 10 billion tokens.}
\label{fig:nsp_mlm}
\end{figure*}

\subsection{Understanding NSP}
To understand the role of NSP in BERT pre-training we compare the performance of three models: the first trained on MLM; the second trained on MLM and NSP; and the third trained on MLM with NSP's context split, but without NSP's loss, which we label split. Contrasting the MLM model to the split model explicates the impact of splitting the inputs context, while comparing the NSP model to the split model clarifies the benefits of the NSP loss. As expected, figure \ref{fig:nsp_mlm} a) demonstrates a clear performance drop when splitting contexts. From figure \ref{fig:nsp_mlm} b) and \ref{fig:nsp_mlm} c), we see the biggest drops are from inference tasks. We hypothesize that providing a model split contexts and no signal to differentiate it from contiguous text hinders it's ability to understand the logical flow of language. As we contrast the NSP model and the split model, we see that adding such a signal does indeed improve the results on inference tasks, especially in early stages of training. However, as training progresses, its benefit stagnate. This may be because, as other papers have proposed, NSP is semantically shallow and can often be solved easily through lexical overlap. Interestingly, figure \ref{fig:nsp_mlm} c) shows that the NSP loss provides no benefits, and may indeed be detrimental towards non-inference tasks even when compared to the split model. Finally, we see that the MLM model continues to improve throughout each stage of training, whereas both the NSP model and the split model see have diminishing returns with more training, indicating that splitting the context imposes inherent limitations on language models.

\subsection{Auxiliary Tasks}
We first compare the 14 auxiliary tasks in Table \ref{table:aux_tasks} to a MLM baseline (No Aux.). As noted in the previous section, and supporting many recent papers \citep{roberta, xlnet, spanbert}, NSP is detrimental to training. As discussed by \citet{psp} and reinforced by the results of \citep{structBERT}, next sentence prediction provides a shallow supervision signal, and is often solvable through lexical overlap. Adjacent sentence prediction and sentence ordering on the other hand require deeper semantic understanding of the structure of language. Our results clearly support this claim, with SO and ASP outperforming MLM and NSP on all inference tasks and greatly outperforming all auxiliary tasks on RTE, the only low-resource inference task. Additionally, they are less penalized by context splitting, which we have shown to degrade performance; NSP and SDP cut the context in half 50\% of the time, ASP cuts the context in half a third of the time, and sentence ordering preserves the full context in all cases, albeit shuffled. The model trained using the Quick Thoughts variant (QT) performs the best out of all the above models. We hypothesize that the loss, based on cosine similarity, provides a soft clustering around semantically similar topics, which produces more distinguishable embeddings. The FastSent variant (FS) provides a similar signal and performs the second best, suggesting that some form of soft clustering does provide substantial benefit to pre-training. TF-IDF, and to a lesser extent TF, prediction also improve performance on a range of downstram tasks. This aligns with \citet{ernie2}'s observations that identifying high value words (and discounting filler words) provides a useful signal for language models. All other tasks fail to provide any meaningful gains. Of these, the context distortion from the corruption prediction tasks (TC and SC) likely outweigh their benefit. Additionally, MLM is already a form of corruption, making TC and SC partially redundant. Our results did not find the Sentence Boundary Objective (SBO) to be beneficial. However, as it was originally implemented for spans, this does not discount the results of \citet{spanbert}; in our context, which only masks a single word, it is likely redundant with MLM. The trigram shuffling (TGS) tasks similarly did not provide the value exhibited in \citet{structBERT}. However, due to a lack of details and code in the original paper, implementation details may be at fault. Token length and capitalization prediction, which were implemented as other proxies for word importance prediction, appear to be too noisy for their intended purpose.

\begin{table*}[ht] 
\centering
\small
\begin{tabular}{|l|c|c|c|c|c|c|c|c|c|c|}
\hline
 & MNLI & QQP & QNLI & SST-2 & CoLA & STS-B & MRPC & RTE & Avg. & Dev. Set Avg.\\ \hline
BERT\textsubscript{Base} & \underline{84.6/83.4} & 71.2 & 90.5 & \underline{93.5} & 52.1 & \underline{85.8} & \underline{88.9} & 66.4 & 79.6 & - \\
MLM baseline & 81.8/81.3 & 70.0 & 87.1 & 90.4 & 45.3 & 80.6 & 87.3 & 59.2 & 76.1 & 80.0 \\ 
CMTL+ & 83.8/82.9 & \underline{71.7} & \underline{90.7} & 92.2 & \underline{56.3} & 83.4 & 88.8 & \underline{66.9} & \underline{80.1}  & \underline{83.2} \\\hline \hline
BERT\textsubscript{Large} (330M) & 86.7/85.9 & 72.1 & 92.7 & 94.9 & 60.5 & 86.5 & 89.3 & 70.1 & 82.1 & - \\
RoBERTa (355M) & 90.8/90.2 & 74.3 & 95.4 & 96.7 & 67.8 & 91.9 & 92.3 & 88.2 & 87.88 & - \\
T5 (11B) & 92.2/91.9 & 75.1 & 96.9 & 97.5 & 71.6 & 92.8 & 92.8 & 92.8 & 89.78 & - \\ \hline

\end{tabular}
\caption{GLUE test results, scored by the evaluation server excluding the problematic WNLI task. Matched/mismatched accuracy are reported for MNLI, F1 scores are reported for QQP and MRPC, Spearman correlations are reported for STS-B, and accuracy scores are reported for the other tasks. The BERT\textsubscript{Base} results are from the original BERT paper \citep{bert}. The MLM baseline and CMTL+ models are our implementations. We include the performance of our models on the development set for reproducibility.  Best results in each column for models of comparable size are underlined. For context, we additionally include results from the GLUE leaderboard for BERT\textsubscript{Large}, RoBERTa, and T5, and their respective size measured by number of parameters. BERT\textsubscript{Base}, MLM baseline, and CMTL+ all have a size of 110M parameters.}
\label{table:glue}
\end{table*}

\begin{table*}[ht]
\centering
\small
\begin{tabular}{|l|c|c|c|c|c|c|c|c|c|c|}
\hline
 & BoolQ & CB & COPA & MultiRC & ReCoRD & RTE & WiC & WSC & Avg. \\ \hline
MLM baseline & 69.2 & 67.8 & 59.0 & 30.7 & 33.0 & 59.4 & 61.0 & 65.1 & 55.7 \\
CMTL+ & \underline{72.0} & \underline{72.9} & \underline{62.8} & \underline{34.9} & 33.0 & \underline{64.0} & \underline{64.9} & 65.1 & \underline{58.7} \\\hline \hline
Bert\textsubscript{Large} (330M) & 77.4 & 79.5 & 70.6 & 47.1 & 71.7 & 71.7 & 69.6 & 65.1 & 69.0 \\
T5 (11B) & 91.2 & 95.4 & 94.8 & 75.7 & 93.8 & 92.5 & 76.9 & 93.8 & 89.3 \\ \hline

\end{tabular}
\caption{SuperGLUE test results, scored by the evaluation server. Both models use most common class prediction for ReCoRD and WSC. The MLM baseline also uses most common class prediction for MultiRC. Best results in each column for models of comparable size are underlined. For context, we additionally include results from the SuperGLUE leaderboard for BERT\textsubscript{Large} and T5, and their respective size measured by number of parameters. CMTL+ and MLM baseline both have sizes of 110M parameters.}
\label{table:superglue}
\end{table*}

\subsection{Combining Tasks}

To test combining multiple tasks, we use all auxiliary losses that substantially outperform a pure MLM baseline. For tasks that provide similar signals, we select the one that achieved a higher average on the previous test; QT is chosen over FS and TF-IDF is chosen over TF. Between ASP and SO, which have a statistically insignificant difference, we choose SO as it retains the full context, as any split context from ASP would likely be detrimental to the other tasks. This provides 4 tasks for the multi-task training: MLM, QT, SO, and TF-IDF. 

Table \ref{table:comb_tasks} shows a stark contrast between incorporating an MLM loss term in each iteration compared to treating MLM equivalent to other tasks when switching between them; Alt.+ and CMTL+ both outperform their counterparts by 1.9 and 3.3 percent respectively. Our results indicate that multi-task training with MLM preserves the benefits of each individual task, with the combined models retaining QT's high CoLA score and SO's high RTE score. Further, these gains are additive in most cases: for QNLI, MNLI, and STS-B the combined models performs better than any single auxiliary task models. This leads to a model that vastly outperforms the baseline MLM model or using any singular auxiliary task. 

Between combination methods that use MLM in every iteration, the incremental approach appears to be the worse, while summing everything, alternating auxiliary tasks (Alt.+), and continual multi-task learning on auxiliary tasks (CMTL+) all perform similarly, with CMTL+ slightly outperforming the other two, which supports \citet{ernie2}'s results. Interestingly, both approaches where tasks vary each iteration (Alt.+ and CMTL+) see a significant benefit on the CoLA task. While not beneficial in our framework, an alternating pattern or CMTL have the additional benefit of enabling different input structures or the use of different corpora (such as ERNIE 2.0's IR relevance corpora), which cannot be done using a direct summation.

\subsection{Final Results}
For our final test, we train our baseline MLM model and CMTL+ model on 32 billion tokens and present the results using the GLUE and SuperGLUE evaluation servers in Tables \ref{table:glue} and \ref{table:superglue} respectively. When fine-tuning these models, we run an exhaustive hyper parameter search on learning rates = 1e-5, 2e-5, 3e-5, 5e-5, batch sizes = 16, 32, and number of epochs = 2, 3, 4. The results show that the CMTL+ model -- trained on MLM, QT, SO, and TF-IDF in a continual multi-task learning framework -- vastly outperforms the MLM baseline in every task. Further, our model trained on 32 billion tokens outperforms the original BERT\textsubscript{Base}, which required 137 billion tokens. While we include larger models -- BERT\textsubscript{Large}, RoBERTa, and T5 -- in the tables for context, we remind the readers that these results are not comparable to our results. First, they are larger, with sizes of 330 million, 335 million, and 11 billion parameters respectively, compared to our 110 million parameters. Second, RoBERTa and T5 are trained using a much larger dataset of 160 GB and 750 GB compared to our (and BERT\textsubscript{Large}'s) 13 GB. Finally, BERT\textsubscript{Large}, RoBERTa, and T5 are trained on more tokens, training on 137 billion, 2.2 trillion, and 1 trillion tokens respectively compared to our 32 billion tokens. While the results are not comparable, we hope that the tasks we used in our model can be utilized by newer and larger models to improve their understanding of language.

\section{Discussion}
Our results support several recent papers: we support \citet{roberta, xlnet, spanbert}'s claim that NSP hinders BERT pre-training, especially for non-inference tasks, due to cutting context half the time;  we reinforce \citet{psp, structBERT}'s proposal that NSP prediction is a semantically shallow and often solvable through lexical overlap and that using a task that requires understanding the ordering of contiguous text provides a stronger semantic signal; and we uphold \citet{ernie, ernie2}'s idea that a language model should be trained in a multi-task setting. Further, we offer novel methods and insights. Providing a signal to reduce the embedding distance between semantically similar sentences, as in our FastSent or QuickThought variants do, produces a strong boosts to downstream tasks, with the hypothesis that they produce more distinguishable embeddings. Providing a signal that relays word importance, such as TF-IDF and TF, likewise produces substantial benefit to BERT pre-training. We show strong evidence that a MLM variant loss should always be included when multi-task learning. Finally, we demonstrate the value of multi-task learning for language model pre-training; combining multiple beneficial tasks leads to better results than using any of the individual tasks alone.  

As our focus was a breadth-based search of possible auxiliary tasks, we believe that further gains are possible through a deeper exploration of each task. Using soft labels in ASP for sentences that are near (but not directly adjacent to) the other sentence has been shown to provide improvements \citep{psp}. ($n!$)-way classification with $n$ sentence-pieces for sentence ordering is a more challenging task that could provide additional benefits. Other similarity metrics, such as dot product or Euclidean distance, may provide more useful for the FS or QT methods. Beyond using a loss based on a similarity metric, it is possible that other unsupervised clustering algorithms could be beneficial. Currently, each task has different loss ranges based on the nature, and not the inherent value, of the task. As some tasks may be more useful than others, it is likely that weighting each task based on some value metric could prove beneficial. Groups with sufficient computational resources may also be interested in exploring how the ordering in the continual multi-task learning framework affects downstream tasks. Lastly, we do not tune hyperparameters, using only the stated values from previous papers for all our experiments. We leave the above potential to future work.

\section{Conclusion}
We investigate and support several reasons why next-sentence prediction is ill-suited for BERT pre-training, we provide better inference-based alternatives, and we develop other novel auxiliary tasks based on word importance and soft clustering that provide substantial benefits to BERT pre-training. We also demonstrate the benefit of multi-task learning in BERT pre-training, and identify key factors on how to best combine tasks. We hope the insights provided here will help guide the development of better language models in the future. 

\section*{Acknowledgements}
Rudzicz is supported by a CIFAR Chair in Artificial Intelligence.

\newpage

\bibliography{olfmlm}
\bibliographystyle{acl_natbib}

\newpage

\appendix

\section{Continual Multi-Task Learning}
\label{sec:cmtl}
Though no explicit details are provided for ERNIE 2.0's continual multi-task learning (CMTL), we infer the following algorithm from their description and example:
\begin{enumerate}
    \item Split training into N stages where N is the number of non-jointly trained tasks (e.g. in our final model, only auxiliary tasks are counted, MLM is not counted). Each stage defines a number of tokens to be trained on for each task. When each task has been trained on for the specified number of tokens, the training moves to the next stage.
    \item Calculate the token chunk size, $C = T / (N * (N + 1))$, where T is the total number of training tokens.
    \item Each stage, $S_i$, a new task is introduced. During that stage the new task is trained on $C * (i + 1)$ tokens, previously introduced tasks are trained on $C$ tokens, and yet to be introduced tasks are trained on $0$ tokens.
\end{enumerate}

The method can use iterations or tokens. The above method trains on each task using the same number of tokens/iterations, gradually incorporating more tasks, while still training on previous tasks. Below we provide two examples. The first from \cite{ernie2} which uses four tasks and 200k iterations, the second from our final combined model which uses three tasks (MLM not included) and 10 billion tokens.

\begin{table}[h!]
\centering
 \begin{tabular}{|l | c | c | c | c |} 
 \hline
 Task & Stage 1 & Stage 2 & Stage 3 & Stage 4 \\ [0.5ex] 
 \hline\hline
 1  & 20k & 10k & 10k & 10k \\ 
 2 & 0 & 30k & 10k & 10k \\
 3 & 0 & 0 & 40k & 10k \\
 4 & 0 & 0 & 0 & 50k \\
 \hline
 \end{tabular}
 \caption{Training using CMTL with 4 tasks over 200k total iterations. Example from \citet{ernie2}}
\label{table:ernie}
\end{table}

\begin{table}[h!]
\centering
 \begin{tabular}{|l | c | c | c | } 
 \hline
 Task & Stage 1 & Stage 2 & Stage 3 \\ [0.5ex] 
 \hline\hline
 1  & 1.67B & 0.83B & 0.83B \\ 
 2 & 0 & 2.5B & 0.83B \\
 3 & 0 & 0 & 3.33B \\ \hline
 \end{tabular}
 \caption{Training using CMTL with 3 tasks over 10B total iterations. }
 \label{table:me_final}
\end{table}

\section{Significance testing}

\label{sec:sig test}
To further solidify our claims, we perform significance testing on our results between NSP and the MLM baseline, as well as our CMTL+ model and the MLM baseline. For each NSP, MLM, and the CMTL+ model we evaluate 5 runs, found in table \ref{table:sig_test}. We first run a Lilliefors test, and find that the p-values are large enough that we accept the null hypothesis that our the data follows a normal distribution for each of our sets of experiments. We then run an independent t-test between NSP and MLM, and between the our CMTL+ and MLM. We correct the p-values using Bonferroni correction and find a p-val of $2.547e-03$ between NSP and MLM and a p-val of $1.069e-06$ between CMTL+ and MLM. In both cases, the p-values are small enough that we reject the null hypothesis that the samples come from the same distribution, supporting our hypothesis that MLM is better than NSP, and CMTL+ is better than MLM.

\begin{table}[h!]
\begin{tabular}{|l | c | c | c |} 
\hline
Run & MLM & NSP & CMTL+ \\\hline
1 & 78.18 & 77.663 & 80.56 \\
2 & 77.90 & 77.363 & 80.30 \\
3 & 78.38 & 77.775 & 80.66 \\
4 & 78.25 & 77.275 & 80.45 \\
5 & 77.96 & 77.338 & 81.03 \\\hline
Mean: & 78.13 & 77.483 & 80.60 \\\hline
Std. Dev.: & 0.20 & 0.22 & 0.27 \\\hline
Lilliefors p-val & 0.712 & 0.148 & 0.659 \\\hline
\end{tabular}
\caption{Average GLUE score results on 5 different trainings.}
\label{table:sig_test}
\end{table}

\end{document}